\documentclass[conference]{IEEEtran}
\IEEEoverridecommandlockouts
\usepackage{cite}
\usepackage{amsmath,amssymb,amsfonts}
\usepackage{algorithmic}
\usepackage{graphicx}
\usepackage{textcomp}
\usepackage{xcolor}
\def\BibTeX{{\rm B\kern-.05em{\sc i\kern-.025em b}\kern-.08em
    T\kern-.1667em\lower.7ex\hbox{E}\kern-.125emX}}

\usepackage{algorithmic}
\usepackage{comment}
\usepackage{blindtext}
\usepackage{url}
\usepackage[ruled,vlined]{algorithm2e}
\usepackage{subcaption}
\usepackage{amssymb}
\usepackage{enumitem}
\usepackage{xcolor}
\usepackage{multirow}
\usepackage{balance}

\DeclareMathOperator*{\argmax}{arg\,max}
\DeclareMathOperator*{\argmin}{arg\,min}

\begin{document}

\title{Byzantine-robust Federated Learning through Spatial-temporal Analysis of Local Model Updates}

\author{
    \IEEEauthorblockN{Zhuohang Li\IEEEauthorrefmark{1}, Luyang Liu\IEEEauthorrefmark{2}, Jiaxin Zhang\IEEEauthorrefmark{3}, and Jian Liu\IEEEauthorrefmark{1}}
    \IEEEauthorblockA{\IEEEauthorrefmark{1}University of Tennessee, Knoxville, TN, USA}
    \IEEEauthorblockA{\IEEEauthorrefmark{2}Google Research, Mountain View, CA, USA}
    \IEEEauthorblockA{\IEEEauthorrefmark{3}Oak Ridge National Laboratory, Oak Ridge, TN, USA}
    \IEEEauthorblockA{Email: zli96@vols.utk.edu, luyangliu@google.com, zhangj@ornl.gov, jliu@utk.edu}
    
}

\maketitle

\begin{abstract}
Federated Learning (FL) enables multiple distributed clients (e.g., mobile devices) to collaboratively train a centralized model while keeping the training data locally on the client\textcolor{black}{s' devices}.
Compared to traditional centralized machine learning, FL offers many favorable features such as offloading operations which would usually be performed by a central server and reducing risks of serious privacy leakage.
However, Byzantine clients that send incorrect or disruptive updates due to system failures or adversarial attacks may disturb the joint learning process, consequently degrading the performance of the resulting model. In this paper, we propose to mitigate these failures and attacks from a spatial-temporal perspective. Specifically, we use a clustering-based method to detect and exclude incorrect updates by leveraging their geometric properties in the parameter space. Moreover, to further handle malicious clients with time-varying behaviors, we propose to adaptively adjust the learning rate according to momentum-based update speculation. Extensive experiments on $4$ public datasets demonstrate that our algorithm achieves enhanced robustness comparing to existing methods under both cross-silo and cross-device FL settings with faulty/malicious clients.
\end{abstract}

\begin{IEEEkeywords}
federated learning, Byzantine robustness, aggregation rule
\end{IEEEkeywords}

\section{Introduction}

The rapid growth of machine learning model complexity and demand for large training datasets has stimulated the interest in distributing the learning task across multiple machines. As an emerging distributed learning paradigm, \textit{federated learning} (FL)~\cite{mcmahan2017communication} allows multiple clients (e.g., mobile devices) to collaboratively learn a shared model in a privacy-preserving way. In contrast to conventional machine learning methods that require all training data to be exposed to a central server, FL allows privacy-sensitive data to be retained on each client. In particular, each client computes an update of the model on their local dataset, and a central server (e.g., the service provider) coordinates the learning process by aggregating the clients' updates to maintain a global model. This strong privacy guarantee of FL has spurred a broad spectrum of real-world applications in areas like mobile computing~\cite{hard2018federated} and telemedicine~\cite{sheller2020federated}.

Despite its favorable characteristics, FL still faces challenges from non-malicious failures (e.g., noisy data) as well as adversarial attacks (e.g., Byzantine attacks~\cite{blanchard2017machine} and backdoor attacks~\cite{bagdasaryan2020backdoor,sun2019can}).
Moreover, the strong emphasis on clients' privacy prevents the server from accessing and inspecting the clients' data directly, which makes detecting these failures and attacks a challenging task~\cite{kairouz2019advances}. The  \textit{aggregation rule} adopted by the central server acts as the most crucial component in ensuring the amount of robustness of FL systems. By default, the server aggregates the local model updates by taking the average value as the global model update~\cite{mcmahan2017communication}. However, it has been shown that 
a single faulty/malicious client can impede the convergence of the jointly learned model under this setting ~\cite{blanchard2017machine}, \textcolor{black}{posing a serious risk to the security of such systems.}

Recently, several theoretical approaches based on gradient similarity~\cite{blanchard2017machine} or robust statistics~\cite{chen2017distributed,yin2018byzantine} have been proposed to achieve Byzantine-resilient learning. Although offering provable guarantees, in practice these methods only provide a weak level of tolerance to attacks and the resulting model could still be significantly influenced by malicious clients. 
To address this, \textit{Bulyan}~\cite{mhamdi2018hidden} proposes to execute another robust aggregation rule for multiple iterations to provide a stricter convergence guarantee at the cost of high computational burden. 
Other methods attempt to detect and remove malicious clients by estimating each client's reliability through calculating the descendant of the loss function~\cite{xie2019zeno} or projecting the clients' updates into a latent space using a variational autoencoder~\cite{li2020learning}. However, these methods require prior knowledge on the clients' data distributions for loss descendent estimation or autoencoder training, which is hard to satisfy in practice, especially for cross-device FL where clients' data are private and extremely heterogeneous.
To defeat time-coupled attacks, existing methods relying on historical data have been proposed to adaptively estimate the quality of client updates using a hidden Markov model~\cite{munoz2019byzantine} or reduce the variance of benign gradients to expose malicious clients via distributed momentum~\cite{el2020distributed,karimireddy2021learning}. However, \textcolor{black}{these methods assume a simple cross-silo scenario with a fixed set of clients that continuously participate the learning process at every communication round, while ignoring the more dynamic cross-device scenario where clients may withdraw or rejoin the FL at any time.} \textcolor{black}{Moreover,} in order to keep track of and update the reliability score of each client, \textcolor{black}{some methods (e.g., \cite{munoz2019byzantine}) require} the server \textcolor{black}{to keep track of} the mapping between the submitted updates and the clients' identities, which may lead to serious risks of privacy breaches (e.g., data inference~\cite{wang2019beyond}, property inference~\cite{ganju2018property,melis2019exploiting}, or membership inference~\cite{nasr2019comprehensive}).

In this work, we seek to relax these constraints by proposing a new aggregation strategy that can resist strong adversarial attacks for achieving Byzantine-resilient federated learning. Different from existing studies, we propose to examine the local model updates from both spatial and temporal perspectives. 
From the spatial perspective, we show that at each round of communication, the updates from faulty/malicious clients exhibit certain distinguishable geometric patterns in the parameter space. Leveraging this observation, we can assess the integrity of a client's model update by inspecting its cosine similarity with all updates and utilize a clustering-based approach to detect and filter out malicious updates. Moreover, to handle malicious clients with time-varying behavior, we propose to adaptively adjust the learning rate at each communication round by comparing the received updates with the speculated update according to historical data from a temporal perspective. This enables our method to tolerant abrupt and uncertain adversarial activities in cross-device FL setting with highly unreliable clients.
\textcolor{black}{Different from existing methods,} our method does not rely on the prior knowledge of the client's data distribution or clients' identities, and therefore can be applied along with existing techniques such as secure shuffling~\cite{bittau2017prochlo} and differential privacy~\cite{geyer2017differentially} to ensure user's privacy.

We conduct extensive experiments to evaluate the proposed method on four public datasets \textcolor{black}{under two realistic federated learning settings: (1) The \textit{cross-silo FL}, which involves a fixed set of clients that continuously participate in the learning process; and (2) The \textit{cross-device FL}, where the participating clients are dynamically selected and changing at each communication round.
Moreover, to investigate its robustness against more advanced attacks, we also evaluate the proposed method against two state-of-the-art time-coupled attacks~\cite{baruch2019little,xie2020fall}.}
The results demonstrate that our method achieves greater robustness in the presence of noisy, faulty or malicious clients comparing to the current state-of-the-art aggregation methods such as Krum~\cite{blanchard2017machine}, Median~\cite{yin2018byzantine}, and Trimmed Mean~\cite{yin2018byzantine}.

\section{Background and Related Work}

\begin{figure}[t]
\centering
\includegraphics[width=0.54\linewidth]{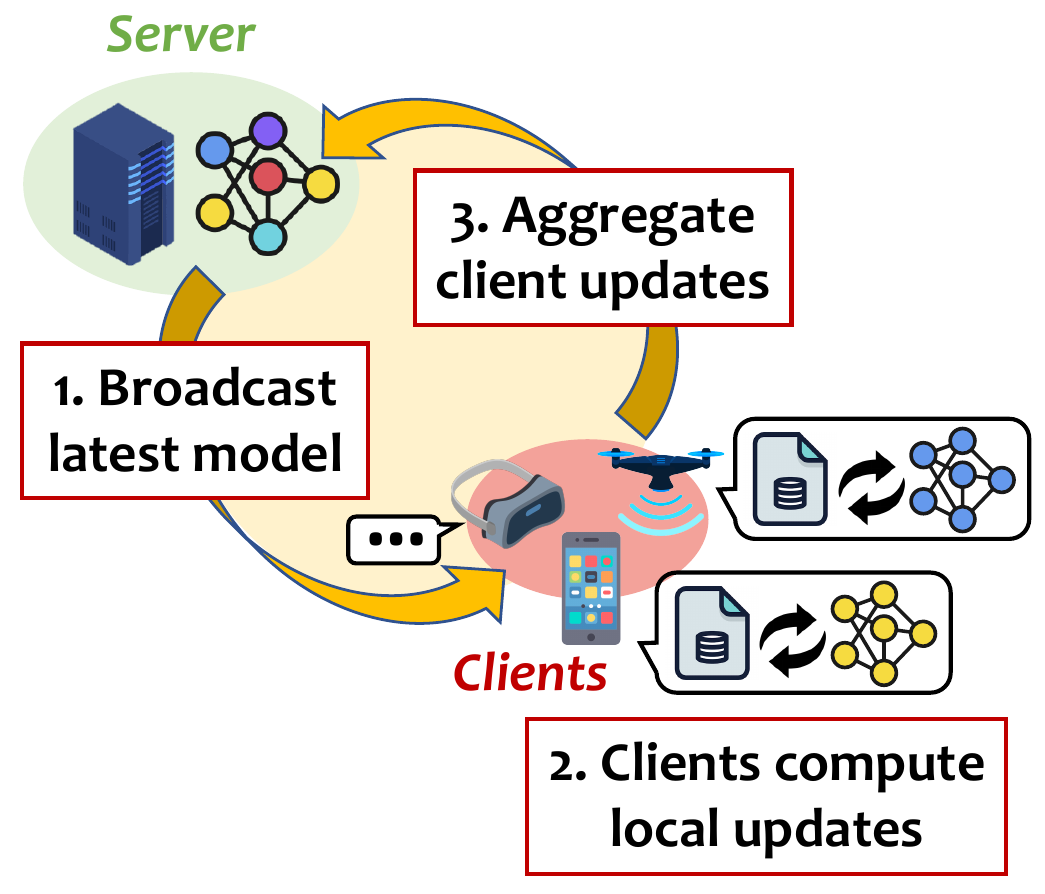}
\vspace{-1mm}
\caption{Illustration of the federated learning framework.}
\vspace{-3mm}
\label{fig:overview}
\end{figure}

\subsection{Federated Learning}
\textit{Federated Learning (FL)} (or \textit{Collaborative Learning}) is a distributed learning framework that allows multiple clients to collaboratively train a machine learning model under the coordination of a central server, while keeping their private training data locally on the device without being shared or revealed to the server or other clients. Federated learning can be conducted among a small set of reliable clients (\textit{cross-silo}) or among a large number of mobile and edge devices (\textit{cross-device}). Let $\mathcal{C}$ denote the set of participating clients, each of which holds a local dataset $\mathcal{D}_k, k \in \mathcal{C}$ of $n_k$ data samples. $\mathcal{D}=\bigcup_{k\in\mathcal{C}}\mathcal{D}_k$ is the joint training dataset and $N=\sum_{k\in\mathcal{C}} n_k$ is the total number of data samples. $\mathcal{L}(w, \mathcal{D}_k)$ represents the empirical loss over a model $w\in \mathbb{R}^d$ and dataset $\mathcal{D}_k$. The objective of federated learning can be formulated as:
\begin{equation}
    \min_{w \in \mathbb{R}^d} \left \{ \mathcal{L}(w, \mathcal{D})=\sum_{k \in \mathcal{C}} \frac{n_k}{N} \mathcal{L}(w, \mathcal{D}_k) \right \}.
\end{equation}
Initially, the central server randomly initializes a global model $w_0$. Then at each communication round, the following steps are performed to achieve the learning objective, as shown in Figure~\ref{fig:overview}:
\begin{itemize}[leftmargin=*]
    \item \textbf{Step I: Broadcast Latest Model.} The central server broadcasts the latest global model $w_t$ to all the clients (usually in cross-silo FL) or a subset of clients ($\mathcal{C}_t$) that are selected to participate in this round of training (usually in cross-device FL).
    \item \textbf{Step II: Clients Compute Local Updates.} Each client computes an update of the model on its local dataset by performing several iterations of gradient descent: $w^{k}_{t+1} \leftarrow w^{k}_{t+1} - \eta \nabla_{w} \mathcal{L}(w_k, \mathcal{D}_k)$, with $\eta$ being the learning rate.
    
    \item \textbf{Step III: Aggregate Client Updates.} The server updates the global model by aggregating the local updates according to a certain aggregation rule $\mathcal{A}(\cdot)$: $w_{t+1} \leftarrow \mathcal{A}(\{w^{k}_{t+1}:k\in\mathcal{C}_t\})$.
\end{itemize}

\subsection{Byzantine-resilient Aggregation Rules}

The most widely-used aggregation rule for communication-efficient FL is Federated Averaging (\textit{FedAvg})~\cite{mcmahan2017communication}, which aggregates the client updates by computing a weighted average: $w_{t+1} \leftarrow \sum_{k\in \mathcal{C}_t} \frac{n_k}{N}w_{t+1}^k$. However, \textit{FedAvg} is not fault-tolerant and even a single faulty/malicious client can prevent the global model from converging~\cite{blanchard2017machine,yin2018byzantine}.
To address this, several robust aggregation techniques have been proposed:

\textbf{Krum~\cite{blanchard2017machine}.} At each communication round, Krum selects $m$ of the $|\mathcal{C}_t|$ local model updates for computing the global model update by comparing the similarity between the provided local updates. Suppose $f$ out of $|\mathcal{C}_t|$ clients are malicious, Krum assigns a score for each local model update $w^k$ by computing the sum of Euclidean distances between $w^k$ and $|\mathcal{C}_t|-f-2$ neighboring local updates that are closest to $w^k$. The $m$ local model updates with the smallest scores will be selected and the average will be computed as the global model update.

\textbf{Median~\cite{yin2018byzantine}.} Median is a coordinate-wise aggregation rule that considers each model parameter independently. Specifically, to decide the $i$th parameter of the global model update, the server sorts the $i$th parameter of the submitted $|\mathcal{C}_t|$ local model updates and takes the median value. Median aggregation can achieve order-optimal statistical error rate if the loss function is strongly convex.

\textbf{Trimmed Mean~\cite{yin2018byzantine}.} Trimmed Mean is another coordinate-wise aggregation rule. At each round of communication, given a trim rate $\gamma$ ($\gamma \in (0, \frac{1}{2})$), the server first sorts the $i$th parameter of the submitted $|\mathcal{C}_t|$ local model updates, removes the smallest and largest $\gamma |\mathcal{C}_t|$ values , and then computes the mean of the remaining $(1-2\gamma)|\mathcal{C}_t|$ values as the $i$th parameter of the global model update. It is proven that trimmed mean can achieve order-optimal error rate for strongly convex losses if $a \leq \gamma < \frac{1}{2}$, where $a=\frac{f}{|\mathcal{C}_t|}$ is the ratio between the number of byzantine clients over the total number of clients.

\begin{figure}[t]
\centering
\includegraphics[width=\linewidth]{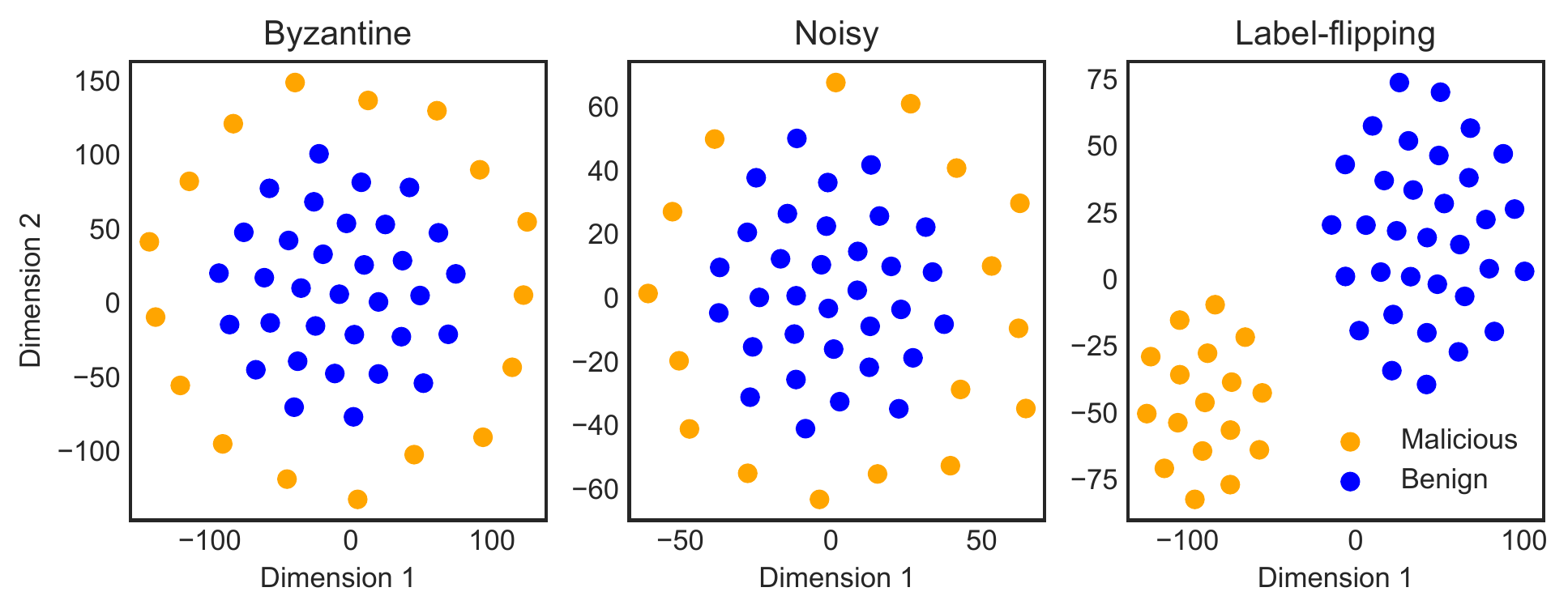}
% \vspace{-3mm}
\caption{Visualization of client updates.}
% \vspace{-3mm}
\label{fig:geo}
\end{figure}

\textbf{Other methods.} \textit{Bulyan}~\cite{mhamdi2018hidden} iteratively executes another byzantine-resilient aggregation rule (e.g., Krum) multiple times to achieve enhanced robustness, but is not scalable due to high computational cost. \textit{Zeno}~\cite{xie2019zeno} computes the descendant score for each update and only aggregates the top $|\mathcal{C}_t| - b$ updates with the highest scores, where $|\mathcal{C}_t|$ is the total number of clients and $b$ is a hyperparameter that needs to be specified in advance and should be no less than the number of malicious clients. A more recent study~\cite{li2020learning} proposes to use a variational autoencoder to project client updates into a latent space where malicious updates can be detected. However, this method is based on the assumption that the server has access to data that are drawn from the same distribution as the client's private data to train the autoencoder, which is hard to satisfy in practice.
Other studies aim to achieve robust federated learning by identifying and blocking the malicious clients through adaptive model quality estimation~\cite{munoz2019byzantine} or clustered federated learning~\cite{sattler2020byzantine}.
However, these methods require the server to keep track of the identity of each client to maintain a trustworthiness score or to establish the cluster structure and therefore cannot be applied to the scenarios where privacy-preserving techniques (e.g., secure shuffling~\cite{bittau2017prochlo}) are applied.
\textcolor{black}{Additionally, methods based on distributed momentum~\cite{ganju2018property,melis2019exploiting} have been proposed to defeat time-coupled attacks that aim to stealthily diverge the model by accumulating small perturbations over time. Despite their effectiveness in overcoming time-coupled perturbations, only the simple cross-silo scenario has been considered, leaving the more dynamic and realistic cross-device scenario unexplored.}

Differently, in this work, we aim to design an aggregation scheme that can tolerant attacks or failures in a more dynamic FL scenario while achieving privacy preservation, i.e., without requiring prior knowledge on the number of faulty/malicious clients, the distribution of the client's data, or the mapping between the submitted model updates and the clients' identities.
Moreover, different from existing Byzantine-resilient aggregators (e.g., Krum, Median, and Trimmed Mean), our method can tolerant stronger attacks that have large negative impact on the joint model with few malicious clients, such as targeted data poisoning attacks~\cite{tolpegin2020data} \textcolor{black}{and time-coupled attacks~\cite{baruch2019little,xie2020fall}}.

\vspace{-0.5mm}
\section{Methodology}
Our algorithm inspects the client updates from two critical perspectives. (1) \textit{Spatial perspective}: We leverage geometric patterns to filter out malicious updates within each round of communication; and (2) \textit{Temporal perspective}: We utilize historical data from previous communication rounds to detect temporal outliers.

\subsection{Spatial Perspective}
\subsubsection{Geometric Property of Malicious Updates}
We first perform a preliminary study to compare the distributions of the model updates computed by benign clients and the updates from faulty/malicious clients. We simulate a simple federated learning task with $50$ clients, \textcolor{black}{$34\%$} of which are either faulty clients that contain \textit{noisy} data or malicious clients that perform \textit{Byzantine} or \textit{label-flipping} attack (detailed settings are described in Section~\ref{sec:setup}). The learning objective is to jointly train a simple multi-layer perceptron model with one hidden layer of $200$ neurons on the MNIST dataset~\cite{lecun1998mnist}. We let each client perform $5$ iterations of gradient descent with a learning rate of $0.01$ on its local dataset and report the model update. Figure~\ref{fig:geo} shows the visualization of the clients' updates \textcolor{black}{selected from an arbitrary communication round} in a 2-dimensional space using t-SNE~\cite{maaten2008visualizing}. From the plots we can observe that these malicious updates diverge from benign updates, causing the aggregated global update to be biased and deviate from the direction of the true gradient, which in turn results in degraded performance of the learned model. However, on the other hand, the divergent model updates produce identifiable patterns that can potentially be utilized for detecting and removing these anomalous model updates to improve the robustness of the aggregation rule.

\subsubsection{Clustering-based Anomalous Update Detection}
Motivated by the geometric property of the malicious updates, we thus propose to adopt a clustering-based method for achieving unsupervised anomalous model update detection. Since it has been shown that different underlying data distribution of clients can be distinguished by inspecting the cosine similarity between their model updates~\cite{sattler2020clustered}, we use cosine similarity as the metric for computing the affinity matrix. Different from conventional clustered federated learning framework~\cite{sattler2020clustered,sattler2020byzantine}, we construct clusters per each communication round and the cluster structure is not carried over to the consecutive rounds after each partition. This disentangles the mapping between the model update and the client's identity to prevent data inference attacks~\cite{wang2019beyond} and ensures that our method is scalable to cross-device scenario with a large crowd of clients. Specifically, at each communication round $t$, we first construct the affinity matrix $S$ prior to the aggregation by computing the pairwise cosine similarities  between the different clients' updates:
\begin{equation}
    S\leftarrow[s_{i,j}],\: s_{i,j}\leftarrow \frac{<\Delta w_{t+1}^i, \Delta w_{t+1}^j>}{\left \| \Delta w_{t+1}^i \right \| \left \| \Delta w_{t+1}^j \right \|} \:(\forall i, j \in \mathcal{C}_t), 
\end{equation}
where $\Delta w_{t+1}^i = w_t - w_{t+1}^i$.
We then apply agglomerative clustering with complete linkage~\cite{mullner2011modern} to partition the clients' updates into clusters of singleton nodes and iteratively merge the currently most closest pair of clusters into a new cluster, until there are only two candidate clusters left:
\begin{equation}
    c_1, c_2 \leftarrow \argmin_{c_1 \cup c_2=\mathcal{C}}(\min_{i\in c_1, j\in c_2} s_{i, j}).
\end{equation}
Then we compute the largest similarity between the two candidate clusters as the criterion for partitioning:
\begin{equation}
    s(c_1, c_2)\leftarrow \max_{i\in c_1, j\in c_2} s_{i, j}.
\end{equation}
The partition process will be proceeded if $s(c_1, c_2)$ is less than a preset threshold $s_t \in (-1, 1)$. Based on the assumption that the majority of clients are not faulty/malicious, we consider the larger cluster of the two as the benign cluster $c$. If $s(c_1, c_2)\geq s_t$, we consider all client updates in this round to be benign. We aggregate the updates that are decided to be benign according to a certain aggregation rule $\mathcal{A}(\cdot)$:
\begin{equation}
    w \leftarrow \mathcal{A}(\{w_{t+1}^k: k \in c\}).
\end{equation}
In our experiment, we choose to use Median as the default aggregation rule for the proposed algorithm as it does not require prior knowledge on the quantity of malicious clients.
The subsequent operations will only be performed on the aggregated benign updates until the next communication round when a new clustering structure is formed.

\subsection{Temporal Perspective}
Different from cross-silo FL where the clients are almost always available, in cross-device FL scenario, the participating clients are usually a large number of mobile or edge devices that are highly unreliable due to their varying battery, usage, or network conditions.
To ensure training speed and avoid impacting the user of the device, the server usually only selects a fraction of clients that are available for computing the global update at each communication round. As a result, the number of faulty clients selected in each communication round is dynamic and highly variable.
In addition, a client may continue to send genuine updates until some point in the learning process when it is compromised by an adversary. Thus solely relying on spatial patterns is insufficient, especially when facing a sudden violent perturbation. 
\subsubsection{Adaptive Learning Rate Adjustment via Momentum-based Update Speculation}
To cope with these time-varying behaviors and achieve temporal robustness, we propose to assess the quality of the aggregated update by comparing it with a speculated value of update that is predicted according to historical statistics. The intuition is that if the current update significantly deviates from previous results, this can indicate an abrupt change in the state of the participating clients (e.g., in extreme case all clients involved in the current round are malicious).

To make a speculation of the update using historical data, we take inspiration from momentum\textcolor{black}{~\cite{polyak1964some}}, which utilizes the past gradients to smooth out the current update to achieve fast and stable convergence. Specifically, we first estimate the gradient using the aggregated updates: $\Delta w \leftarrow w_t - w$. Then we compute an exponential moving average of the gradient according to:
\begin{equation}
    v \leftarrow \beta \cdot v + (1-\beta)\cdot\Delta w,
\end{equation}
where $\beta$ is the decay factor, and $v$ can be seen as a speculated value of the gradient from past updates. The cosine similarity $\alpha$ between the gradient $\Delta w$ and the averaged value $v$ can be obtained. If $\alpha \leq 0$, all updates in the current round will be discarded. Otherwise, we update the global model according to
\begin{equation}
    w_{t+1} \leftarrow w_t - \eta\cdot v,
\end{equation}
where $\eta$ is the learning rate which is adaptively adjusted according to $\alpha$ based on the initial learning rate $\eta_0$: $\eta=\alpha\cdot\eta_0$.
This indicates that our algorithm will take a small step if $v$ and $\Delta w$ disagrees.
A complete procedure of our algorithm is described in Algorithm~\ref{alg:algorithm}.

\begin{algorithm}[tb]
\caption{Robust Aggregation via Spatial-temporal Pattern Analysis}
\label{alg:algorithm}
\textbf{Input}: Client updates $\{w_{t+1}^k:k \in \mathcal{C}_t\}$, global model $w_{t}$, aggregation rule $\mathcal{A}(\cdot)$\\
\textbf{Parameter}: Clustering threshold $s_{t}$, initial learning rate $\eta_0$, momentum $\beta$\\
\textbf{Output}: Updated global model $w_{t+1}$
\begin{algorithmic}[1] %[1] enables line numbers
\STATE $S\leftarrow[s_{i,j}],\: s_{i,j}\leftarrow \frac{<\Delta w_{t+1}^i, \Delta w_{t+1}^j>}{\left \| \Delta w_{t+1}^i \right \| \left \| \Delta w_{t+1}^j \right \|}\:(\forall i, j \in \mathcal{C}_t)$
\STATE $c_1, c_2 \leftarrow \argmin_{c_1 \cup c_2=\mathcal{C}}(\min_{i\in c_1, j\in c_2} s_{i, j})$
\STATE $s(c_1, c_2)\leftarrow \max_{i\in c_1, j\in c_2} s_{i, j}$
\IF {$s(c_1, c_2)<s_t$}
\STATE $c\leftarrow \argmax_{c\in {c_1, c_2}}(|c|)$
\ELSE
\STATE $c \leftarrow c_1 \cup c_2$
\ENDIF
\STATE $w \leftarrow \mathcal{A}(\{w_{t+1}^k: k \in c\})$
\STATE $\Delta w \leftarrow w_t - w$
\STATE $v \leftarrow \beta \cdot v + (1-\beta)\cdot\Delta w$
\STATE $\alpha \leftarrow \frac{<\Delta w, v>}{\left \| \Delta w \right \| \left \| v \right \|}$
\IF {$\alpha\leq 0$}
\STATE $w_{t+1} \leftarrow w_t$
\ELSE
% \STATE $w_{t+1} \leftarrow w_t - \eta\cdot\alpha\cdot v$
\STATE $w_{t+1} \leftarrow w_t - \eta\cdot v, \: \eta=\eta_0\cdot\alpha$
\ENDIF
\STATE \textbf{return} $w_{t+1}$
\end{algorithmic}
\end{algorithm}

\section{Experiments}

\begin{figure*}[t]
\centering
\includegraphics[width=\linewidth]{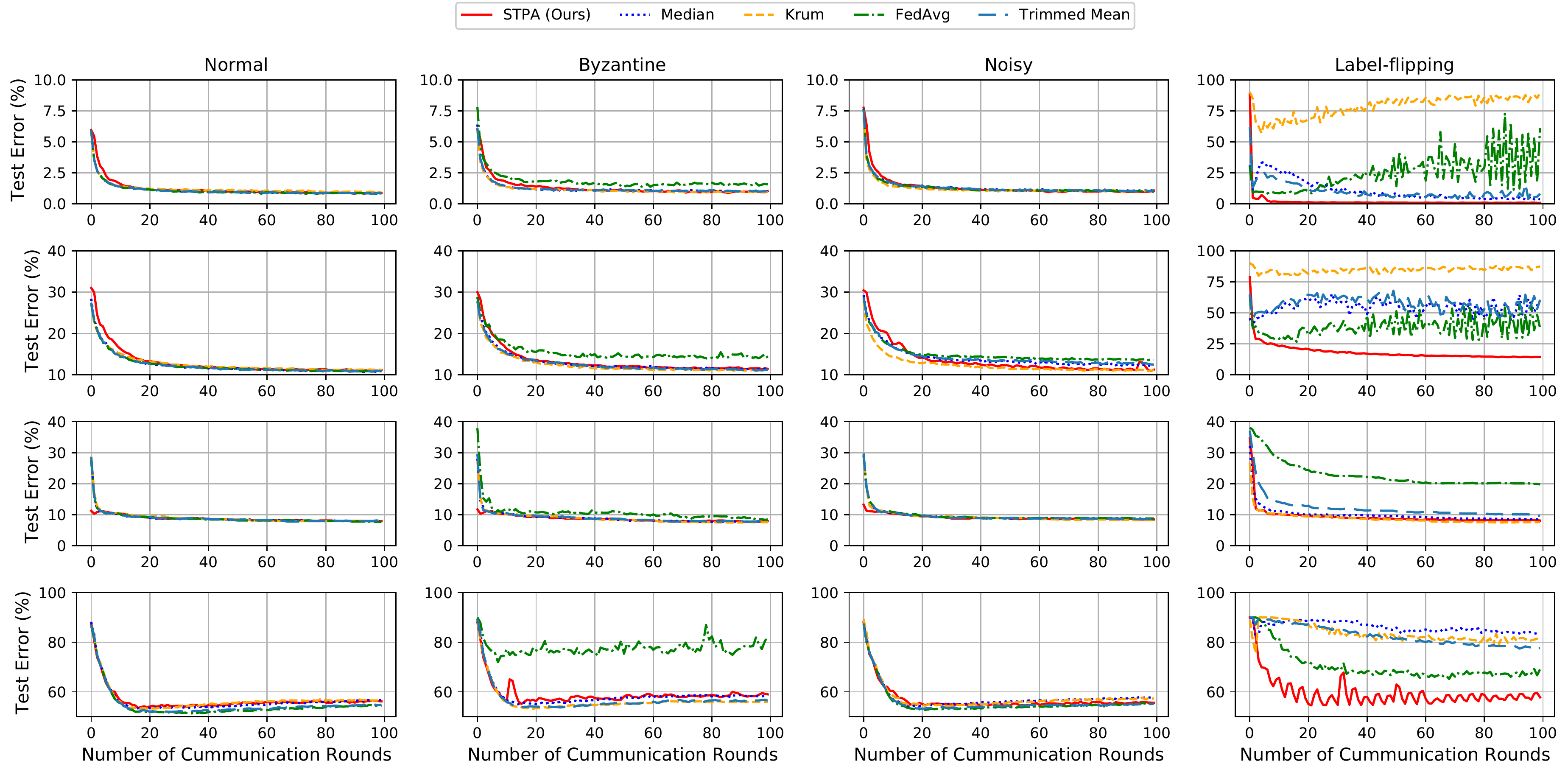}
% \vspace{-3mm}
\caption{Comparison of the baseline methods with the proposed STPA aggregation in the \textit{cross-silo} FL scenario under $4$ different settings. From top to bottom, each row shows the result of the MNIST, Fashion, Spambase, and CIFAR-10 datasets, respectively.}
% \vspace{-4mm}
\label{fig:silo}
\end{figure*}

\subsection{Experimental Setup}\label{sec:setup}
\subsubsection{Federated Learning Scenarios}
Real-world FL systems in production are usually optimized to keep the number of faulty/malicious clients at a low level ($<$$1\%$) using a variety of system-level protections. 
In this paper, in order to better show the superiority of the proposed algorithm, unless mentioned otherwise, we consider an extreme case where around \textcolor{black}{$34\%$} of clients are malicious.
More specifically, we consider the following two federated learning scenarios:

\begin{itemize}[leftmargin=*]
\item \textbf{Cross-silo FL:} There are $20$ clients that continuously participate in every round of communication. We assume that $7$ of them are faulty/malicious. This simulates the federated learning scenario that involves a small number of reliable clients such as different organizations.

\item \textbf{Cross-device FL:} We assume that a total number of $100$ clients are involved and $34$ of which are faulty/malicious. At each round of communication, only $20$ clients are selected randomly to compute the model update. This simulates the federated learning scenario which involves a large number of mobile and edge devices that are unreliable due to varying battery or network conditions.

\end{itemize}

\subsubsection{Baseline Aggregations and Parameter Selection}
In each FL scenario, we compare our proposed spatial-temporal pattern analysis (STPA) algorithm with FedAvg and $3$ representative baseline methods: Krum~\cite{blanchard2017machine}, Median~\cite{yin2018byzantine}, and Trimmed Mean~\cite{yin2018byzantine}.
For fair comparison, we carefully choose the parameters for baseline methods:
for Krum, we assume the number of Byzantine updates $(f)$ is known to the server and set $m$ to be within the range of $[1, |\mathcal{C}_t|-f-2]$ to be Byzantine-resilient; for Trimmed Mean, we set the trim ratio to be within \textcolor{black}{$5\%$-$34\%$}, which is the percentage of the simulated Byzantine clients over total clients.
For our STPA algorithm, we set the $s_t$ to $0.02$, $\beta$ to $0.5$,
\textcolor{black}{and $\eta_0$ to be within the range of $[1.0, 1.6]$.}
% and $\eta_0$ to $1.2$ for the label-flipping setting and $1.6$ for other settings.

\begin{figure*}[t]
\centering
\includegraphics[width=\linewidth]{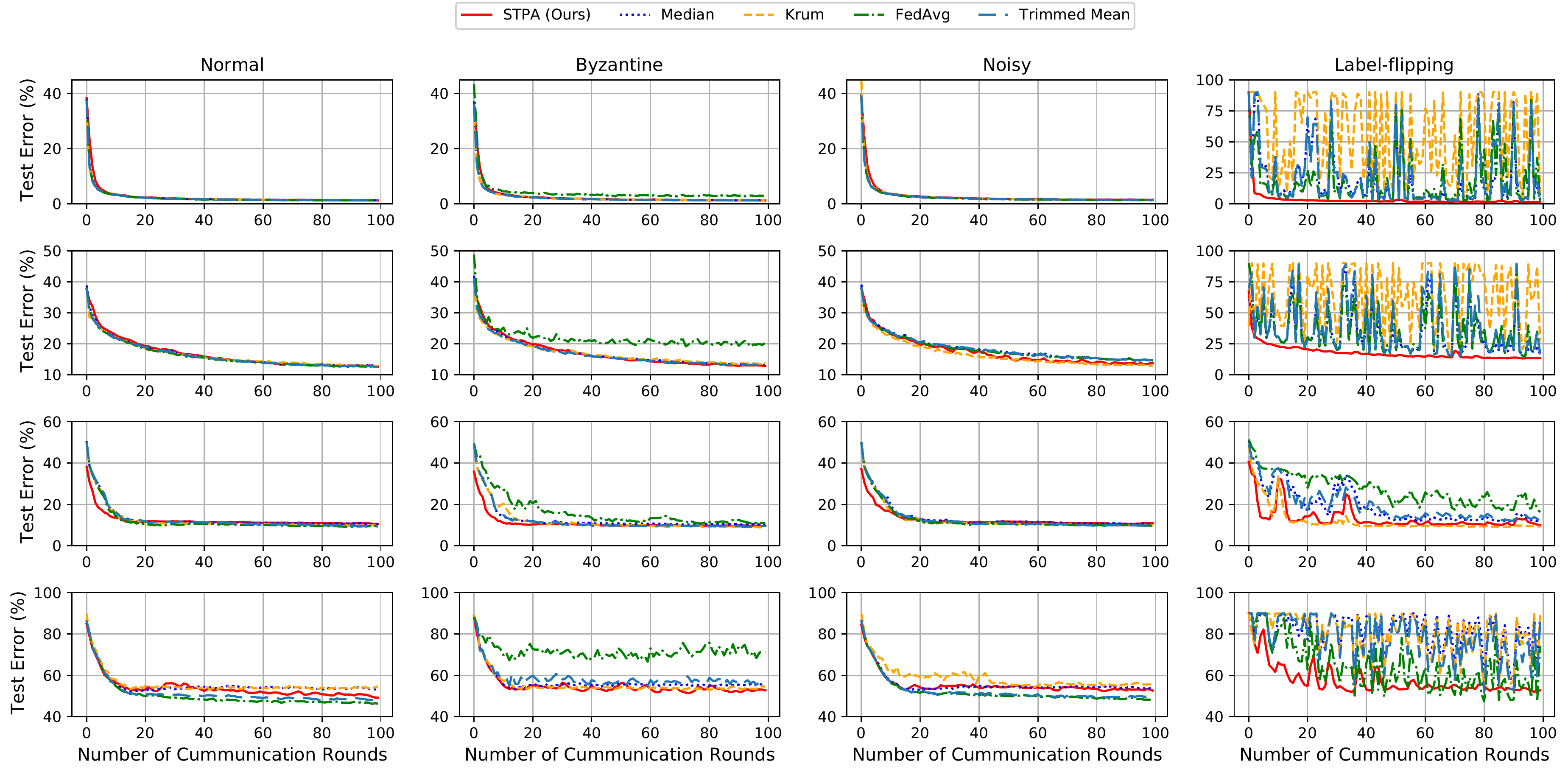}
% \vspace{-3mm}
\caption{Comparison of the baseline methods with the proposed STPA aggregation in the \textit{cross-device} FL scenario under $4$ different settings. From top to bottom, each row shows the result of the MNIST, Fashion, Spambase, and CIFAR-10 datasets, respectively.}
% \vspace{-2mm}
\label{fig:device}
\end{figure*}

\subsubsection{Datasets and Models}
We conduct our experiments on $4$ public datasets: MNIST~\cite{lecun1998mnist}, Fashion-MNIST (Fashion)~\cite{xiao2017fashion},  Spambase~\cite{hopkins1999spambase}, and CIFAR-10~\cite{krizhevsky2009learning}. The MNIST and Fashion-MNIST datasets both contain $70,000$ $28\times28$ gray-scale images from $10$ classes, $60,000$ of which are used for training and the rest are used for testing. The Spambase dataset is a binary classification problem with $4,601$ instances to decide whether an email is spam or not. We keep the first $54$ attributes which indicate whether a particular word was frequently occurring in the e-mail. The dataset is randomly split into training and test sets with a ratio of $8$ to $2$. The CIFAR-10 dataset contains $60,000$ $32\times32$ color images from $10$ classes, with $50,000$ of them being used for training and the rest for testing. For the MNIST and Fashion dataset, we train a convolutional neural network (CNN) with $2$ convolutional layers and $2$ fully-connected (FC) layers.
For Spambase, we train a simple Logistic regression (LR) model.
For CIFAR-10, we train a CNN with $2$ convolutional layer, $1$ max-pooling layer, and $3$ FC layers. A summary of the dataset and model configurations is presented in Table~\ref{tab:dataset}.

\begin{table}[t]
    \centering
    \resizebox{\linewidth}{!}{
    \begin{tabular}{cccccc}
    \hline
    Dataset  & \# Train & \# Test & \# Feature & \# Class & Model \\ \hline
    MNIST    & 60,000   & 10,000  & 784        & 10       & CNN   \\
    Fashion  & 60,000   & 10,000  & 784        & 10       & CNN   \\
    Spambase & 3,680    & 921     & 54         & 2        & LR   \\
    CIFAR-10 & 50,000   & 10,000  & 3,072      & 10       & CNN   \\ \hline
    \end{tabular}
    }
    % \vspace{-1mm}
    \caption{Summary of benchmark datasets and models.}
    % \vspace{-3mm}
    \label{tab:dataset}
\end{table}

\subsubsection{Adversary Model}
For each scenario, we consider the following settings in the experiments: 
\begin{itemize}[leftmargin=*]
    \item \textbf{Normal.} In each communication round, all selected clients perform $5$ steps of gradient descent on their local datasets at a learning rate of $0.01$, and report the genuine local update to the central server.
    
    \item \textbf{Byzantine.} Byzantine clients send model updates that are significantly different from genuine clients. In our experiment, instead of performing gradient descent on their local datasets, the faulty/malicious clients compute model updates drawn from a Gaussian distribution with $0$ mean and isotropic covariance matrix with a standard deviation of $20$.
    
    \item \textbf{Noisy.} For the MNIST, Fashion, and CIFAR-10 datasets, prior to the training procedure, we normalize the image data to $[-1, 1]$. When computing update, a uniform noise is added to the data of the selected noisy clients: $x \leftarrow x + u, u \in U(-1.4, 1.4)$ and $x$ is then clipped to the $[-1, 1]$ interval. For the Spambase dataset, a uniform noise $u \in U(-0.5, 1.5)$ is added to the noisy clients, and the value is then clipped to the $[0, 1]$ interval.
    
    \item \textbf{Label-flipping.}
    All the training labels of the malicious clients are set to zero\textcolor{black}{, which corresponds to the label of the first class}.
    This simulates a strong targeted data poisoning attack scenario, where the adversary's goal is to cause bias in the global model towards a specific class.
\end{itemize}

\textcolor{black}{In addition, we consider the following two state-of-the-art time-coupled attacks that can circumvent existing Byzantine aggregation rules by accumulating small perturbations over many training rounds to eventually diverge the joint model:}
\begin{itemize}[leftmargin=*]
    \item \textcolor{black}{\textbf{Inner Product Manipulation (IPM)~\cite{xie2020fall}.} The goal of this attack is to cause divergence by manipulating the aggregated gradient to deviate from the direction of the true gradient. Suppose $\{g^k: k \in c\}$ is the set of gradients from all benign clients at the current round, the malicious clients in this attack submit Byzantine gradients computed from $-\epsilon \cdot Mean(\{g^k: k \in c\})$.}
    
    \item \textcolor{black}{\textbf{A Little is Enough (ALiE)~\cite{baruch2019little}.} This attack aims to prevent model convergence by hiding small perturbations within the variance of benign gradients. The attack works by controlling the malicious clients to send gradients computed from $Mean(\{g^k: k \in c\})-\epsilon \cdot \sqrt{Var(\{g^k: k \in c\})}$.}
\end{itemize}

\begin{table*}[h]
\centering
\resizebox{0.99\linewidth}{!}{
\begin{tabular}{c|c|ccc|ccc|ccc|ccc}
\hline
\multirow{2}{*}{Scenario}     & \multirow{2}{*}{Method}           & \multicolumn{3}{c|}{5\% Faulty/Malicious Clients}           & \multicolumn{3}{c|}{10\% Faulty/Malicious Clients}          & \multicolumn{3}{c|}{20\% Faulty/Malicious Clients}          & \multicolumn{3}{c}{\textcolor{black}{34\%} Faulty/Malicious Clients}           \\ \cline{3-14} 
                              &                                   & Byzantine & Noisy & Label-flipping & Byzantine & Noisy & Label-flipping & Byzantine & Noisy & Label-flipping & Byzantine & Noisy & Label-flipping \\ \hline
\multirow{5}{*}{Cross-silo}   & FedAvg                            & 2.43$\pm$0.08         & 1.63$\pm$0.04     & 1.53$\pm$0.02              & 3.37$\pm$0.17         & 1.59$\pm$0.02     & 1.68$\pm$0.02              & 5.91$\pm$0.32         & 1.68$\pm$0.01     & 2.84$\pm$0.02              & 1.57$\pm$0.06         & 1.01$\pm$0.02     & 36.10$\pm$20.15              \\
                              & Krum                              & 1.68$\pm$0.04         & 1.61$\pm$0.04     & 1.62$\pm$0.04              & 1.61$\pm$0.04         & 1.62$\pm$0.03     & 1.58$\pm$0.03              & 1.62$\pm$0.03         & 1.53$\pm$0.03     & 19.51$\pm$7.71              & \textbf{0.96$\pm$0.02}         & 1.04$\pm$0.02     & 86.10$\pm$2.10              \\
                              & Median                            & 1.46$\pm$0.04         & 1.60$\pm$0.03     & 1.49$\pm$0.03              & 1.46$\pm$0.03         & 1.43$\pm$0.02     & 1.56$\pm$0.03              & 1.53$\pm$0.05         & 1.64$\pm$0.01     & 1.91$\pm$0.03              & 1.00$\pm$0.02         & 1.08$\pm$0.04     & 4.37$\pm$0.75              \\
                              & \multicolumn{1}{l|}{Trimmed Mean} & \textbf{1.46$\pm$0.01}         & 1.56$\pm$0.05     & \textbf{1.47$\pm$0.04}              & 1.61$\pm$0.03         & 1.73$\pm$0.04     & 1.95$\pm$0.05              & 1.66$\pm$0.07         & 2.03$\pm$0.09     & 2.43$\pm$0.03              & 1.02$\pm$0.03         & 1.06$\pm$0.02     & 7.51$\pm$3.46              \\
                              & STPA (Ours)                       & 1.48$\pm$0.03         & \textbf{1.43$\pm$0.04}     & 1.49$\pm$0.02              & \textbf{1.43$\pm$0.02}         & \textbf{1.39$\pm$0.03}     & \textbf{1.54$\pm$0.03}              & \textbf{1.48$\pm$0.03}         & \textbf{1.49$\pm$0.02}     & \textbf{1.59$\pm$0.03}              & 0.98$\pm$0.02         & \textbf{0.99$\pm$0.03}     & \textbf{0.95$\pm$0.07}              \\ \hline
\multirow{5}{*}{Cross-device} & FedAvg                            & 2.28$\pm$0.09         & 1.49$\pm$0.02     & 1.52$\pm$0.03              & 3.33$\pm$0.23         & 1.46$\pm$0.03     & 1.45$\pm$0.03              & 5.86$\pm$0.47         & 1.56$\pm$0.02     & 2.67$\pm$1.06              & 2.96$\pm$0.10         & 1.42$\pm$0.05     & 14.81$\pm$24.33              \\
                              & Krum                              & 1.49$\pm$0.02         & 1.32$\pm$0.04     & 1.38$\pm$0.04              & \textbf{1.30$\pm$0.03}         & 1.32$\pm$0.03     & \textbf{1.39$\pm$0.04}              & 1.33$\pm$0.04         & \textbf{1.30$\pm$0.02}     & 3.71$\pm$6.81              & 1.38$\pm$0.05         & 1.42$\pm$0.04     & 42.75$\pm$32.74              \\
                              & Median                            & 1.41$\pm$0.04         & 1.34$\pm$0.04     & 1.33$\pm$0.03              & 1.42$\pm$0.05         & 1.33$\pm$0.03     & 1.40$\pm$0.07              & 1.36$\pm$0.04         & 1.45$\pm$0.04     & 1.87$\pm$0.24              & 1.33$\pm$0.03         & 1.42$\pm$0.05     & 11.08$\pm$25.18              \\
                              & \multicolumn{1}{l|}{Trimmed Mean} & \textbf{1.34$\pm$0.03}         & 1.30$\pm$0.03     & 1.31$\pm$0.04              & 1.36$\pm$0.04         & 1.33$\pm$0.03     & 1.42$\pm$0.04              & 1.33$\pm$0.03         & 1.38$\pm$0.03     & 1.81$\pm$0.27              & 1.30$\pm$0.03         & 1.45$\pm$0.04     & 10.05$\pm$21.80              \\
                              & STPA (Ours)                       & 1.38$\pm$0.02         & \textbf{1.26$\pm$0.03}     & \textbf{1.29$\pm$0.06}              & 1.39$\pm$0.17         & \textbf{1.29$\pm$0.03}     & 1.40$\pm$0.03              & \textbf{1.31$\pm$0.02}         & 1.34$\pm$0.04     & \textbf{1.54$\pm$0.04}              & \textbf{1.25$\pm$0.03}         & \textbf{1.41$\pm$0.04}     & \textbf{1.47$\pm$0.06}              \\ \hline
\end{tabular}
}
% \vspace{-1mm}
\caption{Test error ($\%$) with different fractions of faulty/malicious clients on the MNIST dataset.}
% \vspace{-3mm}
\label{tab:num_byz}
\end{table*}

\begin{figure}[t]
    \centering
    \vspace{-3mm}
	\includegraphics[width=0.98\linewidth]{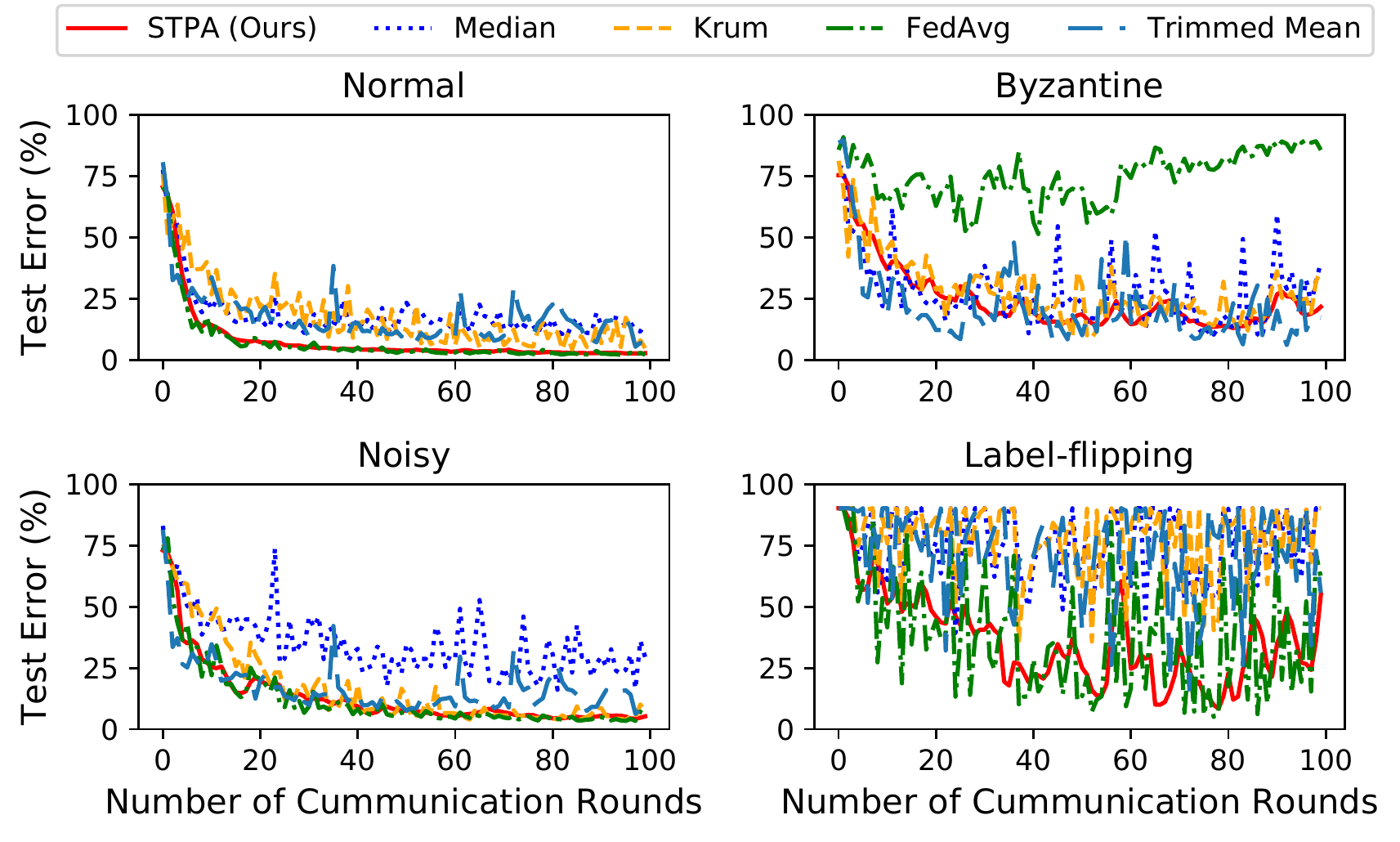}
% 	\vspace{-2mm}
	\caption{Results in the \textit{non-IID} setting on the MNIST dataset.}
% 	\vspace{-2mm}
	\label{fig:noniid}
\end{figure}

\subsection{Experimental Results}

\textbf{Cross-silo FL Scenario.} Figure~\ref{fig:silo} illustrates the experiment results in the cross-silo scenario. From the results, we can observe that the proposed STPA not only achieves comparable convergence speed in the normal setting, but also remains robust in all $3$ faulty/attack settings. Although Krum, Median, and Trimmed Mean are able to achieve satisfactory performance in the Byzantine and noisy settings, they fail to resist the stronger label-flipping attack. Krum has the highest test error in the label-flipping on the MNIST and Fashion datasets ($\sim90\%$) in the cross-silo scenario. This is because Krum selects the most reliable updates by calculating and comparing the Euclidean distance. In the case of label-flipping attack, all malicious updates are biased towards the same class, which increases the probability of being falsely selected by Krum.

\textbf{Cross-device FL Scenario.} The experiment results in the cross-device scenario are shown in Figure~\ref{fig:device}. As we can see, similar to the cross-silo scenario, Median, Krum, and Trimmed Mean are still susceptible to the label-flipping attack in the cross-device scenario. \textcolor{black}{Moreover,} due to the varying number of malicious clients \textcolor{black}{being selected at each round}, the test errors of these methods \textcolor{black}{become} highly variable between \textcolor{black}{consecutive} communication rounds. Our proposed STPA, however, is able to cope with this dynamic scenario by utilizing the temporal patterns to obtain a stabilized gradient and provide \textcolor{black}{enhanced robustness against Byzantine updates}.
\textcolor{black}{Most noticeably in the case of CIFAR-10 dataset, we can achieve the lowest test error against label-flipping attack that is comparable to the performance when there is no attack (i.e., only increased by $3.53\%$), while other methods cannot converge at all.}

\textbf{Fraction of Faulty/Malicious Clients.} To study the impact of the quantity of faulty/Byzantine clients, we vary the fraction of faulty/Byzantine clients from $5\%$ to \textcolor{black}{$34\%$} for both FL scenarios on the MNIST dataset. For each adversary model per scenario, we record the average test errors and their standard deviations of the last $10$ communication rounds. The results are summarized in Table~\ref{tab:num_byz}, with best results being marked as bold. As we can see, our proposed method achieves the lowest test error in $17$ out of $24$ cases when comparing to the baseline methods. In some cases where other methods have better performance, we observe that our test errors are almost at the same level with them (i.e., the difference is typically within $0.1\%$). It is also worth noting that our method can resist the strong label-flipping attack even with a large number of malicious clients whereas all other methods fail. These results further confirm the effectiveness, robustness, and generalization of our proposed aggregation method.

\textbf{Non-IID Setting.}
We conduct experiments in the non-IID federated learning setting where each client is assigned $2$ shards with each containing $300$ image samples from a single class.
Figure~\ref{fig:noniid} shows the results on the MNIST dataset with $100$ clients ($34$ of which are malicious). We observe that all methods show worse performance comparing to the IID setting: FedAvg cannot converge in the Byzantine scenario, Median converges slowly in the noisy scenario, and Krum and Median both fail to converge in the label flipping scenario. Though the proposed algorithm also converges slowly in some scenarios, it can still achieve the stablest convergence curve comparing to other methods.

\begin{figure}[t]
    \centering
    \vspace{-3.8mm}
	\includegraphics[width=0.98\linewidth]{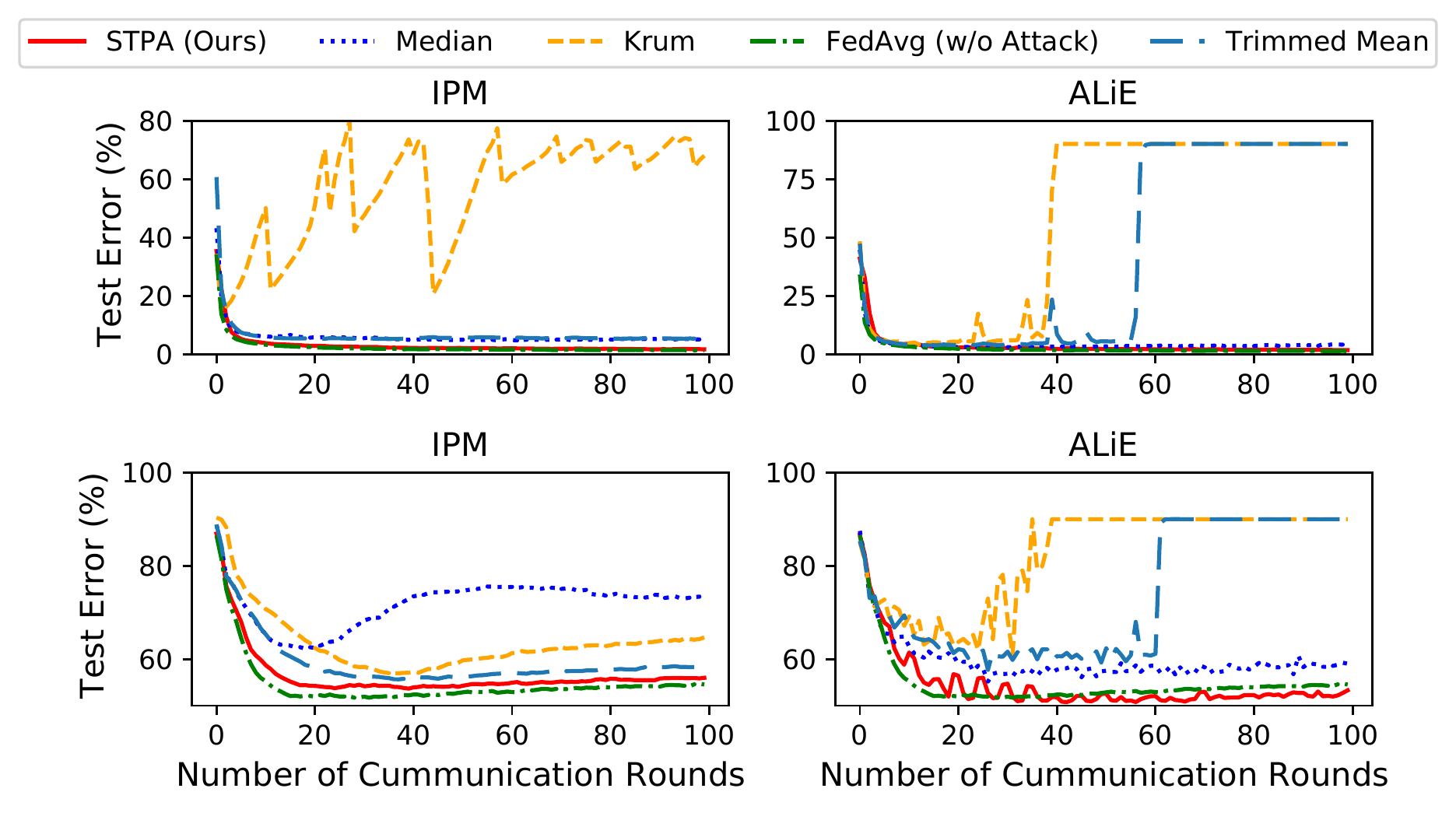}
% 	\vspace{-2mm}
	\caption{\textcolor{black}{Results against \textit{time-coupled attacks}~\cite{baruch2019little,xie2020fall} on the MNIST (1st row) and CIFAR-10 (2nd row) dataset. FedAvg without attack is shown for comparison.}}
% 	\vspace{-2mm}
	\label{fig:time-coupled}
\end{figure}

\textcolor{black}{
\textbf{Time-coupled Attacks.}
To evaluate the robustness against more advanced time-coupled attacks, we conduct experiments on both the MNIST and the CIFAR-10 dataset with $20$ clients ($7$ of which are malicious). For the \textit{Inner Product Manipulation (IPM)}~\cite{xie2020fall} attack, the $\epsilon$ is set to be $1.0$ for the MNIST dataset and $0.3$ for the CIFAR-10 dataset respectively, while the $\epsilon$ of the \textit{A Little is Enough (ALiE)}~\cite{baruch2019little} attack is set to be $1.5$ for both MNIST and CIFAR-10 datasets as is suggested in the original paper. Figure~\ref{fig:time-coupled} presents the results of different aggregation rules against the time-coupled attacks on the two datasets. For each setting, the result of FedAvg without applying any attack is also shown to serve as a reference for demonstrating the effects of the attacks. From the results, we observe that the IPM attack is able to completely break the Krum aggregation rule on the MNIST dataset, while also causing a negative impact on the convergence of other aggregation rules (i.e., increasing error rate by $3\%-4\%$ comparing to FedAvg with no attack.) On the CIFAR-10 dataset, the IPM attack causes the final test errors of Trimmed Mean, Krum, and Median to be increased by $3.75\%$, $10.10\%$, and $18.70\%$, respectively. The ALiE attack is found to be extremely effective against Krum and Trimmed Mean on both datasets, leading to complete model divergence. The ALiE attack also increases the test error of Median by $4.46\%$ on the CIFAR-10 dataset. Our method, on the other hand, is able to resist these time-coupled attacks on both datasets, achieving low test errors that are closest to the ones produced by FedAvg with no attack across all settings.
}

\section{Conclusion}
In this work, we propose a new method to achieve Byzantine-resilient FL through analyzing the spatial-temporal patterns of the clients' updates. By utilizing a clustering-based method, we can detect and exclude incorrect updates in each round of communication. Moreover, to further handle malicious clients with time-varying behaviors, we perform a momentum-based update speculation and adaptive learning rate adjustment. Different from existing methods, our method does not rely on prior knowledge of the client's data distribution or the clients' identities, thereby preserving the user's privacy. We conduct extensive experiments on $4$ public datasets with one normal setting and three faulty/attack settings under both cross-silo and cross-device scenarios. \textcolor{black}{In addition, we also evaluate our method against two state-of-the-art time-coupled attacks.} The results show that our method can achieve enhanced robustness across all settings comparing to the baseline methods.

\bibliographystyle{IEEEtran}
\bibliography{main}

\end{document}